\documentclass[letterpaper]{article} % DO NOT CHANGE THIS
\usepackage{aaai23}  % DO NOT CHANGE THIS
\usepackage{times}  % DO NOT CHANGE THIS
\usepackage{helvet}  % DO NOT CHANGE THIS
\usepackage{courier}  % DO NOT CHANGE THIS
\usepackage[hyphens]{url}  % DO NOT CHANGE THIS
\usepackage{graphicx} % DO NOT CHANGE THIS
\usepackage{natbib}  % DO NOT CHANGE THIS AND DO NOT ADD ANY OPTIONS TO IT
\usepackage{caption} % DO NOT CHANGE THIS AND DO NOT ADD ANY OPTIONS TO IT
\usepackage{amsmath}
\usepackage{amssymb}
\usepackage{algorithm}
\usepackage{algorithmicx}
\usepackage{algpseudocode}
\usepackage{subfigure}
\usepackage{bm}
\usepackage{float}
\usepackage{cases}
\usepackage{enumerate}
\usepackage{multicol}
\usepackage{multirow}
\usepackage{paralist}
\usepackage{amsfonts}

\floatstyle{ruled}
\newfloat{listing}{tb}{lst}{}
\floatname{listing}{Listing}
\title{Entropy Induced Pruning Framework for Convolutional Neural Networks}
\author{
    Yiheng Lu,
    Ziyu Guan,
    Yaming Yang,
    Maoguo Gong,
    Wei Zhao,
    Kaiyuan Feng
}
\affiliations{
}

\begin{document}

\maketitle

\begin{abstract}
Structured pruning techniques have achieved great compression performance on convolutional neural networks for image classification task.
However, the majority of existing methods are weight-oriented, and their pruning results may be unsatisfactory when the original model is trained poorly. That is, a fully-trained model is required to provide useful weight information. This may be time-consuming, and the pruning results are sensitive to the updating process of model parameters.
In this paper, we propose a metric named Average Filter Information Entropy (AFIE) to measure the importance of each filter. It is calculated by three major steps, i.e., low-rank decomposition of the "input-output" matrix of each convolutional layer, normalization of the obtained eigenvalues, and calculation of filter importance based on information entropy. By leveraging the proposed AFIE, the proposed framework is able to yield a stable importance evaluation of each filter no matter whether the original model is trained fully.
We implement our AFIE based on AlexNet, VGG-16, and ResNet-50, and test them on MNIST, CIFAR-10, and ImageNet, respectively. The experimental results are encouraging. We surprisingly observe that for our methods, even when the original model is only trained with one epoch, the importance evaluation of each filter keeps identical to the results when the model is fully-trained. This indicates that the proposed pruning strategy can perform effectively at the beginning stage of the training process for the original model.
\end{abstract}
\section{Introduction}
Convolutional Neural Networks (CNNs) have achieved impressive success on image classification with large model size ~\cite{kriz:ima,christian-szegedy:googlenet,simonyan-et-al:vgg16,hekaiming:resnet}. However, their growing scale imposes huge pressure on computational resources, and may easily lead to the overfitting problems due to the redundancy across the whole network. Therefore, network pruning techniques are studied comprehensively by researchers, to cut off redundant computation branches from original models.

Generally, there are two categories of pruning strategies, i.e., structured pruning~\cite{channelpruning,ThiNet,pavlo:taylor,zhuangliu:networkslimming} and unstructured pruning~\cite{braindamage,brainsurgeon,lottery}. 
Structured pruning is to remove the whole filters from each model layer.
Unstructured pruning is to prune partial parameters from the original model.
Structured pruning has been proved more efficient than unstructured pruning because the structured pruning can eliminate the number of feature maps, which can reduce the consumption of FLOPs apparently \cite{liuzhuang:rethingvalue}. Therefore, in this work, we mainly focus on structured pruning methods.

Structured pruning methods can be further classified into layers-importance-supported (LIS) methods and filters-importance-supported (FIS) methods.
LIS methods aim to evaluate the layer importance at first, and then remove filters according to the layer importance. For example, \cite{ting-wu:layercompensate,suau:pfa,haoli:pruningfilters} utilize different weight-oriented strategies to evaluate the importance of each convolutional layer based on fully-trained models. Differently, FIS methods focus on evaluating the importance of each filter straightly. For example, \cite{pavlo:taylor,zhuangliu:networkslimming,smaller-norm-less-informative} evaluate the importance of each filter by the gradient information and BatchNorm neurons, which can achieve global pruning automatically.
However, both LIS methods and FIS methods above require a fully pre-trained model to provide meaningful parameters. Otherwise, they would yield degenerated pruning results if the original model is not fully-trained.

% \cite{ting-wu:layercompensate} utilized meta-learning to dub filters through layers-compensate. \cite{suau:pfa} proposed PFA-En and PFA-KL to measure the layer importance. \cite{haoli:pruningfilters} used $L_{1}$-norm to stand for the importance of each layer. 
% However, these LIS methods require extra evaluation strategy to determine which filter should be removed, and the importance evaluation of layers and filters both rely heavily on parameters for original model, which may be unconvincing when the pre-defined model is trained poorly. 

% decomposed the pruning task to polynomial with Taylor-expansion, which simplified the evaluation of each filter with the product of weight value and gradient for the output feature. Similarly, \cite{zhuangliu:networkslimming,smaller-norm-less-informative} applied different constraints on BachNorm layers to limit trivial BatchNorm neurons to be zero. Then the filters within the nearest previous convolutional layer would be removed correspondingly.

In this paper, we put forward a entropy based method to evaluate the importance of each filter by the proposed Average Filter Information Entropy (AFIE), which can obtain reliable evaluation for each filter even when the original model is trained with only one epoch. In practice, AFIE can even yield the same evaluation of filters for a fully-trained model and a one-epoch-trained model. 

As shown in Figure \ref{AFIE_figure}, the overall procedure consists of three steps. At first, the "input-output" matrix for a specified convolutional layer is decomposed into low-rank space. Then, the eigenvalues will be constrained by the '0-1' normalization and softmax. Finally, information entropy will be adopted to calculate the AFIE, as well as allocate each layer with different pruning ratio $\lambda$.

Constrained by the complexity of convolutional neural networks, low-rank decomposition of matrix could provide a simple space to analyze the feature of each convolutional layer. Specifically, we use SVD to project the "input-output" tensor of each convolutional layer into a low-rank space, resulting in the left singular matrix, eigenvalues matrix, and right singularly matrix. Then we can treat each eigenvalue as the projection of efficient information along a specified dimension. Namely, the number of eigenvalues stands for the size of dimension, and absolute value of each eigenvalue stands for the strength of projected information along each dimension. 

After the SVD projection, we can obtain a series of decomposed eigenvalue matrices. In order to compare the eigenvalues of each layer with same magnitude, we normalize each eigenvalue matrix to constrain each value between 0 and 1. 

Finally, we introduce the concept of "energy" from Boltzmann machine to describe the importance of decomposed eigenvalue matrix by transferring the absolute value of each eigenvalue into possibilities. As we know, the energy in Boltzmann machine describes how possible one status of a specified neuron will appear, which can be calculated by the softmax operation over possibilities of all status for a specified neuron. We regard each normalized eigenvalue as a possible status of the efficient projection for the specified layer, and adopt softmax to calculate the possibilities of all projection at the low-rank space for the specified layer.

Notably, the sum value of possibilities for all normalized eigenvalues within a layer will equal 1, which means we can describe the projection of a specified layer by the information entropy. After we calculate the total information entropy for a specified layer, the importance of each filter can be specified. Inspired by the conclusion that the structure dominated the representation ability of the network \cite{liuzhuang:rethingvalue}, we assume that \emph{each filter has same importance within one layer}, then we can evaluate the importance of each filter by the Average Filter Information Entropy, which can be calculated by dividing the total information entropy by the number of filters. Consequently, we can authorize the pruning ratio of each layer by the corresponding AFIE of each filter.

We prove that the update of parameters between the one-epoch-trained and fully-trained models is Lipchitz continuous with the gradient of the one-epoch-trained model. It means that we can get similar distribution of parameters for a fully-trained model and a one-epoch-trained model, when the gradient of the one-epoch-model is small. This facilitates our AFIE-based method to yield stable evaluation.     
% Creatively, we define the useful information within a model as knowledge, which can keep unchanged when the structure and datasets of the model is specified. Then the information entropy is utilized to describe the knowledge that is contained across all layers. Therefore, we can use AFIE to evaluate the importance of each layer while without considering the parameters information of original model. Specially, the layers with higher AFIE would contain more knowledge, which may contribute stronger representation power for image classification task. The calculation of AFIE can be formulated through three steps. Initially, we derive the eigenvalues of the 'input-output' matrix for a specified convolutional layer through SVD, as well as normalizing eigenvalues with '0-1'. Then the normalized eigenvalues will be constrained with softmax to make sure the sum value of all eigenvalues are equal to 1. Considering the randomness of parameters updating, we can regard the constrained eigenvalues as the probability of knowledge for the specified layer. Then the knowledge can be described by information entropy, and consequently, we will yield AIFE through the value of information entropy divided by the number of filters within the specified layer. In order to verify the efficiency of proposed algorithm, we apply AFIE on AlexNet, VGG-16, ResNet-50 with MNIST, CIFAR-10 and ImageNet, respectively. Impressively, we achieved comparable pruning with previous pruning algorithms while with obvious lower computation consumption both in training and pruning process on specified network.

In summary, we make three major contributions in this paper.
\begin{itemize}
  \item We observe that previous weight-oriented methods are sensitive to the update of parameters, which can not yield convincing pruning results when the original model is not trained well.
  \item We propose an entropy based method to describe the importance of each filter without considering the weight information. Consequently, we can obtain identical importance evaluation of each filter no matter whether the original model is fully-trained.
  \item We conduct extensive experiments. It turns that for our framework, even when the original model is trained with only one epoch, the pruning results are comparable with that of previous methods.
\end{itemize}
 % The major contribution of this work is that we propose an entropy based method to describe the importance of each filter by the AFIE, which can obtain stable evaluation of filters no matter the original model is trained well. Especially, we can get credible evaluation of AFIE even when the original model is only trained with one epoch. Another contribution is that we achieve comparable results as previous structured pruning methods through AFIE on AlexNet, VGG-16 and ResNet-50 with MNIST, CIFAR-10 and ImageNet, respectively. 

\section{Related Work}
In this section, we briefly introduce the existing Layer-Importance-Supported (LIS) pruning methods and the existing Filters-Importance-Supported (FIS) pruning methods.

\subsection{LIS Pruning Methods}
Study \cite{haoli:pruningfilters} explores the sensitiveness of each convolutional layer through several layer-wise pruning experiments with $\mathit{l}_{1}$-norm. This work stresses the importance of sensitive layers and removes filters from insensitive layers.
Similarly, study \cite{sparse_structure} proposes a coarse-grained pruning method to prune the layers iteratively through sensitivity analysis.
Study \cite{Nonlinear_Reconstruction} achieves layer pruning by minimizing the reconstruction error of nonlinear units. It employs a greedy algorithm to remove trivial neurons.
Study \cite{ting-wu:layercompensate} treats the pruning problems as a unified problem with layer scheduling and ranking problem. It compensates for the approximation error across layers during derivation with the assistance of meta-learning.
Study \cite{suau:pfa} utilizes principal filter analysis to yield compact model by exploiting the intrinsic correlation of filters responses within layers.
Study \cite{NISP} proposes a final layer response mechanism to propagate layer importance from last layer to shallow layer by minimizing the reconstruction error.
All the above-mentioned LIS methods evaluate the importance of layers and filters, and they have achieved impressive pruning results on the image classification task. However, LIS methods usually require a well-pre-trained model to provide useful information for supporting the evaluation.

\subsection{FIS Pruning Methods}
Study \cite{channelpruning} introduces a two-phases channel pruning with LASSO regression, which exploits the redundancy inter feature maps.
Study \cite{ThiNet} proposes the ThiNet to compress models at both training and inference processes at the channel level.
Study \cite{softfilter} uses soft filter technique to remove filters with large model capacity, which can learn more useful information from the training data, as well as reduce the dependence on the pre-trained model.
Study \cite{smaller-norm-less-informative} forces the output of a part of the filters as a constant. Then, the constant is removed by adjusting the bias of impacting layers.
Study \cite{pavlo:taylor} simplifies the pruning process through Taylor expansion with the first order. Then the importance of each filter is evaluated by the products of the weight value and the gradient. 
Study \cite{zhuangliu:networkslimming} regularizes BatchNorm layers, and corresponding filters would be removed if BatchNorm neurons were evaluated with small value. 
Study \cite{metapruning} jointly utilizes meta-learning and evolutionary algorithm to prune a well-trained model automatically. Additionally, some sparsity constraints on filters are also explored comprehensively.
Study \cite{hanzhou:tcn} and study \cite{learningneurons} add extra sparsity on a group of filters, which can remove less important filters by iterative pruning process.
Study \cite{zehao:datadriven} introduces a scale factor with sparsity constraints to select some useful filters within original architecture, which avoid the heavy searching process compared with previous structure sparsity methods.
However, these methods above also require a well pre-trained model to clarify the status of each filter, and the evaluation may be twisted with the updating of parameters from the original model.
% Therefore, we proposed a entropy based method to capture the importance of each convolutional layer at early stage of training process through information entropy, then obtained the importance of each filter by layer entropy. Besides, the entropy based method could assist us to determine the exact pruning ratio of each layer when the total pruning ratio was specified.

\section{Methodology}

\begin{figure*}
    \centering
    \includegraphics[width=\textwidth]{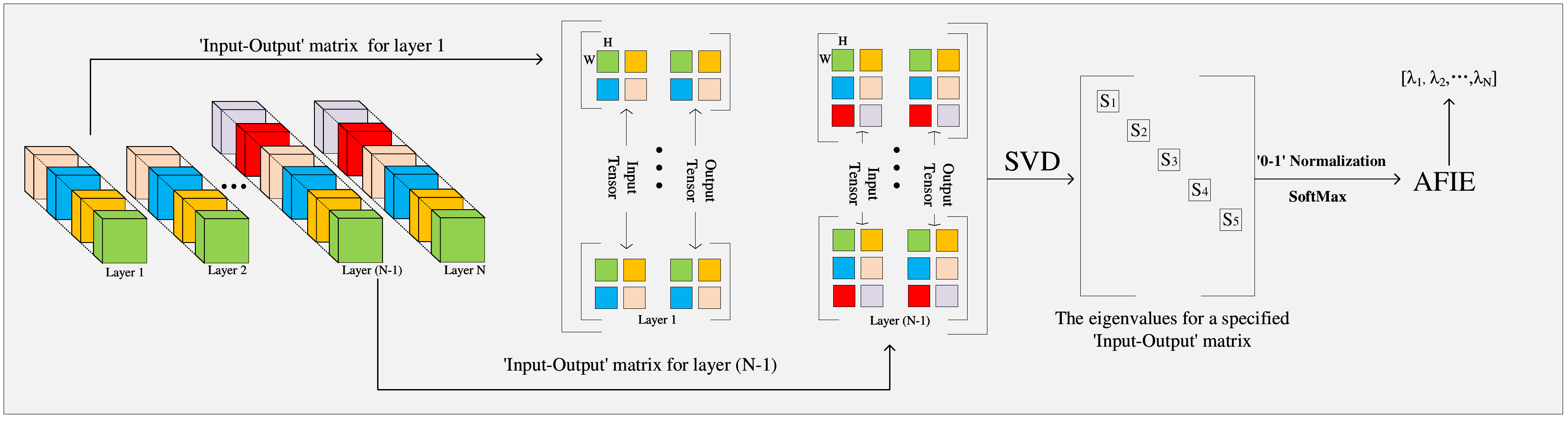}
    \caption{The calculation of AFIE for each convolutional layer. The "input-output" matrix $\textbf{M}\in \mathbb{R}^{\mathcal{I}\times \mathcal{O}\times\mathcal{H}\times{W}}$, where $\mathcal{I}$ and $\mathcal{O}$ are the number of input and output filters, and $\mathcal{H}$ and $\mathcal{W}$ are the height and width of the filters. The $\textbf{M}$ can be decomposed by SVD, and each eigenvalue $s_1, s_2,...,s_5$ will be normalized by '0-1' and softmax to project each eigenvalue as the possibility with the value between 0 and 1, as well as cater for the 1 summation for all normalized eigenvalues within a layer. Finally, we calculate the average filter information entropy for each filter to allocate the different pruning ratio $\lambda_1,\lambda_2,...,\lambda_N$ for each convolutional layer.}
    \label{AFIE_figure}
\end{figure*}

% Motivated by the work from \cite{distribution}, who concluded that the importance of each convolutional layer was related to the distribution of parameters within the specified layer. 
We note that a previous work \cite{distribution} concludes that the importance of each convolutional layer is related to the distribution of parameters in a convolutional layer. Motivated by this work, we quantify the distribution of parameters of each layer from the perspective of information entropy.

The overall idea of the proposed pruning process is described in Figure \ref{AFIE_figure}. The pruning procedure consists of three main steps, i.e., decomposition, normalization, and information entropy calculation. Specifically, the "input-output" matrix $\textbf{M}$ is decomposed by SVD. Then, the obtained eigenvalues are normalized by softmax to map each eigenvalues into possibilities. Finally, the information entropy is calculated based on the obtained possibilities. We can use the obtained information entropy to quantify the filter importance. Intuitively, if the distribution of normalized eigenvalues is flat, then, less redundancy can be found because the projection of original model has similar strength along each dimension.   

We analyze the Lipchitz limitation at the first order for the updating of parameters. It implies good consistency of AFIE for evaluating the importance of each filter between the fully-trained model and the one-epoch-trained model, i.e., we can obtain stable evaluation of AFIE no matter whether the original model is fully-trained.

\subsection{Low-Rank Space Projection}
Usually, the parameters within a convolutional layer is described by a four dimensional "input-output" matrix $\textbf{M}^{(\mathcal{I} \times \mathcal{O} \times \mathcal{H} \times \mathcal{W})}$. It is difficult to straightly analyze the values of $\textbf{M}$ . Therefore, we first fold $\textbf{M}$ as a two dimensional matrix along the $\mathcal{H}$ and $\mathcal{W}$ by averaging the $\mathcal{\mathcal{H}} \times \mathcal{\mathcal{W}}$, denoted as follows:
\begin{align}
    \textbf{M}_{l}^{(\mathcal{I}^{*}\times \mathcal{O}^{*})}=Aver_{(\mathcal{H},\mathcal{W})}(\textbf{M}_{l}^{(\mathcal{I} \times \mathcal{O} \times \mathcal{H} \times \mathcal{W})}),
    \label{average_dimension_reduction}
\end{align}

Usually, SVD can be used for principal components analysis, which yield a eigenvalue matrix to represent the property of original matrix. Here, we adopt SVD to perform the low-rank projection for  $\textbf{M}_{l}^{({\mathcal{I}^{*}\times \mathcal{O}^{*}})}$:
\begin{align}
    (\textbf{U}*\textbf{s}*\textbf{V})_{(l)}=\emph{SVD}(\textbf{M}_{l}^{\mathcal{I}^{*}\times \mathcal{O}^{*}}).
    \label{svd}
\end{align}
where $\textbf{s}$ stands for the decomposed eigenvalues, which is a diagonal matrix that has sorted eigenvalues along the diagonal. $\textbf{U}$ and $\textbf{V}$ are the left singular matrix and right singular matrix, respectively.

Originally, the eigenvalues can be used to describe the zoom level of each axis for the standard orthogonal basis. Here, We aim to use the distribution of eigenvalues to represent the importance of the layers because the eigenvalues maintain the majority information of original matrix. Creatively, we redefine the eigenvalues as the projection of useful information for the specified layer at different dimensions. The number and absolute value of eigenvalues stand for the dimensions and strength of the projection, respectively. Then we can obtain the efficient information of the specified layer only by analyzing the decomposed eigenvalues.

\subsection{Normalization of the Eigenvalues}
In order to perform the comparison between each convolutional layer at same magnitude, we apply '0-1' normalization to the eigenvalues, so as to map each value to the range [0,1]. This process is described as follows:
\begin{align}
    s_{l}^{(i,norm)}=\frac{s_{l}^{i}-s_{l}^{min}}{s_{l}^{max}-s_{l}^{min}},\nonumber \\
    s.t. \quad i=1,2,...,p_l.
    \label{0_1_normalization}
\end{align}
where $s_{l}^{(i,norm)}$ and $s_{l}^{i}$ stand for $i$-th normalized and original eigenvalues for layer $l$, respectively, $s_{l}^{min}$ and $s_{l}^{max}$ stand for the minimum and maximum value within $\textbf{s}_{l}$, respectively, and $p_l$ is the number of eigenvalues for layer $l$. To constraint the summation of all the eigenvalues to 1, we further perform softmax normalization on these eigenvalues, as follows:
\begin{align}
    s_{l}^{(i,soft)}=\frac{ \exp^{s_{l}^{(i,norm)}}}{\sum\limits_{i=1}^{p_l}\exp^{s_{l}^{(i,norm)}}}.
    \label{softmax}
\end{align}

Thus, we can make the parameter distributions of different convolutional layers comparable in the low-rank space. This facilitates the quantification of the layer importance, as described in the next step.

% \begin{algorithm}[htb]
%     \caption{Network Pruning with AFIE}
%     \begin{algorithmic}[1]
%     \Require
%         A hand-craft network with specified dataset:$\mathcal{F}_\mathcal{D}$; A overall pruning ratio: $\lambda^{*}$; The depth of original model: $N$. 
%     \Ensure The measure factor of importance for layer $l$: $AFIE_{l}$; Sparsity ratio of layer $l$: $\lambda_{l}$
%     \State Train $\mathcal{F}_\mathcal{D}$ with little epochs that is less that optimal epochs.
%     \State Decompose the calculation matrix of each convolution and yield the corresponding eigenvalue $s$ according to Equation (\ref{svd}) and (\ref{average_dimension_reduction}).
%     \State Normalize $s$ with Equation (\ref{0_1_normalization}) and Equation (\ref{softmax}).
%     \State $l=0$.
%       \For {$l<N$}
%             \State Derive the measure factor of each layer according to \Statex \quad\; Equation (\ref{total_information_entropy}) and (\ref{AFIE}), then yield $AFIE_{l}$ for layer $l$.
%             \Statex \quad\;\,$l++$.
%         \EndFor
%     \State Formulate the pruning sparsity of each layer with Equation (\ref{sparsity_confirmation}), (\ref{sprasity_calculation}) and (\ref{sparsity_finally}) under the overall pruning ratio $\lambda^{*}$. Then yield $\lambda_{l}$ for layer $l$.
%     \State Apply one-shot pruning with $\lambda_{l}$ on $\mathcal{F}_\mathcal{D}$ to yield the well pruned model.
%     \State Train above well pruned model with fully training. 
%     \end{algorithmic}
%     \label{AFIE_Algorithm}
% \end{algorithm}

\subsection{Average Filter Information Entropy}
Consequently, we borrow the concept of "energy" from Boltzmann machine to transfer each eigenvalue into possibilities, which can describe how possible a projection along one dimension will response. Higher energy means the specified layer contains more useful information along one projected dimension. We can quantify the energy for a specified layer by the information entropy, which can disentangle the importance of each layer in a mathematical way. The computation of information entropy is described as follows:
\begin{align}
    H(x)=-\sum\limits_{x\in \Psi}p(x)\log p(x),
\label{information_entropy}
\end{align}
where $x$ and $p(x)$ stands for a specified status and the corresponding possibility, $\Psi$ is the set of all possible status within a fixed system. For image classification task with CNNs, we regard each convolutional layer as an independent system (item). The decomposed eigenvalues can be used to represent all the possible status (projection) for describing useful information.

We denote the total useful information for layer $l$ as $\emph{K}_{l}$, which is computed as follows:
\begin{align}
    \emph{K}_{l}=-\sum\limits_{i=1}^{p_l}s_{l}^{(i,soft)}\log s_{l}^{(i,soft)}.
    \label{total_information_entropy}
\end{align}

% we denote the status of layer $l$ as $\textbf{s}_{l}^{O(norm)}$ between 0 and 1 with softmax. Then the value of eigenvalues can be used to represent the probability of occurrence for each eigenvalue within layer $l$. the softmax operation can be described as:
% \begin{align}
%     s_{I(l)}^{O(i,soft)}=\frac{ s_{I(l)}^{O(i,norm)}}{\sum\limits_{i=1}^{k_l}s_{I(l)}^{O(i,norm)}},
%     \label{softmax}
% \end{align}
% Then we could write the $\emph{K}_{l}$ as:

Obviously, both $s_{l}^{(i,soft)}$ and $p_{l}$ rig the calculation of $\emph{K}_l$, which means the number of filters will impose huge effect to evaluate the total useful information for a layer. 
We note that a previous work \cite{liuzhuang:rethingvalue} concludes that the representation ability of CNNs is dominated by the structure of the model. Inspired by this conclusion, here, we can reasonably assume that \emph{each filter has the same importance within one layer} because all the filters within a convolutional layer usually have the same structure. Thus, we can obtain the importance of each filter for layer $l$ by dividing the $\emph{K}_{l}$ over the number of filters $c_{l}$, arriving at the proposed AFIE metric for layer $l$:
\begin{align}
    \emph{AFIE}_{l}=\frac{\emph{K}_l}{c_{l}}.
    \label{AFIE}
\end{align}
% where $T_{l}$ is the number of filters in layer $l$. When the $\textit{AFIE}$ of a layer is calculated,

The obtained $\emph{AFIE}_{l}$ describes the importance of each filter for layer $l$, and thus the importance of all the filters across all layers can be quantified.

\subsection{Filters Pruning with AFIE}
After obtaining AFIE of all layers, we allocate the pruning ratio for each layer according to the following equation:
\begin{align}
    \lambda_{l}=\lambda_{min}\frac{\emph{AFIE}_{max}}{\emph{AFIE}_{l}},\nonumber \\
    s.t.\quad \sum\limits_{l}^{\emph{N}}\lambda_{l}p_{l}= \lambda^{*}p^{*},
\label{sprasity_calculation}
\end{align}
where $\lambda_{l}$ stands for the pruning ratio for layer $l$, $\lambda^{*}$ and $p^{*}$ respectively stand for the specified overall pruning ratio and the total number of filters for the original model, and $\emph{N}$ is the number of the convolutional layers. Thus, we can calculate the pruning ratio of each layer by equation (\ref{sprasity_calculation}).

Notably, the layer with the maximum AFIE, i.e. $\emph{AFIE}_{max}$ should be allotted with the minimum pruning ratio, i.e. $\lambda_{min}$. According to Equation (\ref{sprasity_calculation}), we also have the following derivation:
\begin{align}
    \lambda^{*}p^{*}=  \sum\limits_{l}^{\emph{N}}\lambda_{min}&\frac{\emph{AFIE}_{max}}{\emph{AFIE}_{l}}p_{l},\nonumber \\ 
    &\Downarrow \nonumber \\
    \lambda_{min}=\sum\limits_{l}^{\emph{N}}&\frac{\lambda^{*}p^{*}\emph{AFIE}_{l}}{\emph{AFIE}_{max}p_{l}},
    \label{sparsity_finally}
\end{align}

% \begin{align}
%     \sum\limits_{l}^{\emph{N}}\lambda_{l}p_{l}= \lambda^{*}p^{*},
% \label{sparsity_confirmation}
% \end{align}
% where $\lambda_{l}$ stands for the pruning ratio for layer $l$. $\lambda^{*}$ and $p^{*}$ stand for the specified overall pruning ratio and the total number of filters for original model, and $\emph{N}$ is the number of the convolutional layers. Notably, the layer with the highest AFIE should be allotted with minimum pruning ratio, which can be denoted as $\lambda_{min}$. Then the pruning ratio of layer $l$ could be written as:
% \begin{align}
%     \lambda_{l}=\lambda_{min}\frac{\emph{AFIE}_{max}}{\emph{AFIE}_{l}},
% \label{sprasity_calculation}
% \end{align}
% where $\emph{AFIE}_{max}$ stands for the layer with highest importance. 

In particular, we reserve a small set of filters (1\% in this work) within each convolutional layer to maintain the integrity of the topology for original model. Otherwise, the gradient information can not flow from deep layers to shallow layers when a whole layer is removed. Thus, we regulate equation (\ref{sparsity_finally}) as:
\begin{equation}
    \lambda_{l}=
   \begin{cases}
        &\lambda_{min}\frac{\emph{AFIE}_{max}}{\emph{AFIE}_{l}},\quad \lambda_{min}\frac{\emph{ AFIE}_{max}}{\textit{AFIE}_{l}}<1,  \\
        &0.99, \qquad \qquad \quad \,\,\,\, \lambda_{min}\frac{\emph{AFIE}_{max}}{\emph{AFIE}_{l}}\geq 1.
    \end{cases}
 \label{sparsity_regulate}
\end{equation}

Thus, we can determine the pruning ratio of each convolutional layer quickly by solving above equations. This process avoids massive iterative pruning and retraining, saving much computational resources.

\subsection{Theoretical Analysis of AFIE}
We provide the theoretical analysis to show that AFIE can be accurately and consistently computed no matter whether the original model is fully-trained. We use $\mathbf{M}$ and $\textbf{M}^{*}$ to denote the "input-output" matrix of a model that is fully-trained and under-trained (not converged), respectively.
\newtheorem{theorem}{Remark}
\begin{theorem}
The difference between and $\mathbf{M}^{*}$ and $\mathbf{M}$ is sufficiently small under the constraint of Lipschitz limitation, denoted as $\mathbf{M}^{*} \approx_L \mathbf{M}$.
\label{theorem}
\end{theorem}
We prove \textbf{Remark} \ref{theorem} by illustrating that the updating of parameters between the fully-trained and under-trained models is sufficiently small under the constraint of Lipschitz limitation. We denote the filters of the network as $\bm{f}=\{f_{1}^{1},f_{1}^{2},...,f_{\emph{N}}^{c_{\emph{N}}}\}$, where $f_{\emph{N}}^{c_{\emph{N}}}$ stands for the $c_{\emph{N}}$-th filter for layer $N$. The fully-trained and under-trained models are defined as $\mathcal{F}(\mathcal{D},\bm{f})$ and $\mathcal{F}(\mathcal{D},\bm{f}^{*})$, respectively. Thus, we can decompose the fully-trained model by the first-order Taylor expansion when $\bm{f}=\bm{f}^{*}$:
% \begin{align}
%     \mathcal{F}(\mathcal{D},\bm{f})=\mathcal{F}(\mathcal{D},\bm{f}=\bm{f}^{*})&+\frac{\partial \mathcal{F}(\mathcal{D},\bm{h}=\bm{f}^{*})}{\partial(\bm{f}=\bm{f}^{*})}(\bm{f}-\bm{f}^{*})\nonumber \\
%     &\Downarrow \nonumber \\
%     \frac{\mathcal{F}(\mathcal{D},\bm{f})-\mathcal{F}(\mathcal{D},\bm{f}^{*})}{\bm{f}-\bm{f}^{*}}&=\frac{\partial \mathcal{F}(\mathcal{D},\bm{f}^{*})}{\partial\bm{f}^{*}}\nonumber \\
%     &\Downarrow\nonumber\\
%     \nonumber \\
%     Lip(\bm{f}) = &Grad(\bm{f}^{*})\Rightarrow \textbf{M} \approx_L \textbf{M}^{*}\nonumber \\
%     \nonumber \\
%     s.t. \lim \limits_{\bm{f}^{*}}\; Grad(f^{*}) \approx_L 0.
% \end{align}

\begin{align}
    \mathcal{F}(\mathcal{D},\bm{f})=\mathcal{F}(\mathcal{D},\bm{f}=\bm{f}^{*})&+\frac{\partial \mathcal{F}(\mathcal{D},\bm{h}=\bm{f}^{*})}{\partial(\bm{f}=\bm{f}^{*})}(\bm{f}-\bm{f}^{*})\nonumber \\
    &\Downarrow \nonumber \\
    \frac{\mathcal{F}(\mathcal{D},\bm{f})-\mathcal{F}(\mathcal{D},\bm{f}^{*})}{\bm{f}-\bm{f}^{*}}&=\frac{\partial \mathcal{F}(\mathcal{D},\bm{f}^{*})}{\partial\bm{f}^{*}}\nonumber \\
    &\Downarrow\nonumber\\
    \nonumber \\
    Lip(\bm{f}) = &Grad(\bm{f}^{*})\nonumber \\
    &\Downarrow\nonumber\\
    \textbf{M} \approx_L \textbf{M}^{*}, s.t. &\lim \limits_{\bm{f}^{*}}\; Grad(f^{*}) \approx 0.
\end{align}
where $Lip(\cdot)$ and $Grad(\cdot)$ stand for the Lipschitz limitation and the gradient for the specified parameters. The above derivation implies that the updating of parameters between the fully-trained and under-trained models are constrained by the gradient of the under-trained model, i.e., the updating of parameters for the model is Lipschitz continuous with respect to $\frac{\partial \mathcal{F}(\mathcal{D},\bm{f}^{*})}{\partial\bm{f}^{*}}$. Usually, the variation of the gradient is large only at beginning of the training process, and $\frac{\partial \mathcal{F}(\mathcal{D},\bm{f}^{*})}{\partial\bm{f}^{*}}$ decreases dramatically as the training epochs increase. Therefore, we claim that the updating of the parameters is bounded by the gradient of the under-trained model. The gap of the parameters between the fully-trained and under-trained models are small even when the under-trained model is trained with few epochs. 

From $\textbf{M}\approx_L \textbf{M}^{*}$, we have $\textbf{s}_{l}^{fully}\approx_L\textbf{s}_{l}^{non}$. This facilitates the model yield stable AFIE to evaluate the importance of each layer even when the original mode is trained with only few epochs. In the experiments, we show that our method compute consistent AFIE scores for each layer when the original model is only trained with one epoch.

% reduce the possibility to be trapped into local optimization for the evaluation of layer importance. Consequently, one-shot pruning is resorted to prune original model.

% Our pruning framework can be concluded as Algorithm\ref{AFIE_Algorithm}.
% In order to verify the efficiency of AFIE, we implement our pruning framework on AlexNet, VGG-16 and ResNet-50 with MNIST, CIFAR-10 and ImageNet, respectively.

\section{Experiments}

\setlength{\tabcolsep}{1.5mm}
\begin{table}[ht]
\scriptsize
  \centering
  \caption{The evaluation of each convolutional layer for AlexNet on MNIST.}
    \begin{tabular}{|c|c|c|ccccc|}
    \hline
    \multirow{3}[0]{*}{Model} & \multirow{3}[0]{*}{Epochs} & \multirow{3}[0]{*}{Top1-Acc} & \multicolumn{5}{c|}{AFIE} \\
    \cline{4-8}
    & & &$\emph{AFIE}_{1}$&$\emph{AFIE}_{2}$&$\emph{AFIE}_{3}$&$\emph{AFIE}_{4}$&$\emph{AFIE}_{5}$\\
    \hline
    \multirow{3}[0]{*}{AlexNet}&1 & 0.955  & 0.015 & 0.022 & 0.014 & 0.022 & 0.022 \\
    \cline{2-8}
    &10 & 0.988 & 0.015 & 0.022 & 0.014 & 0.022 & 0.022 \\
    \cline{2-8}
    &20 & 0.992 & 0.015 & 0.022 & 0.014 & 0.022 & 0.022 \\
    \hline
    \end{tabular}%
  \label{k_alexnet}%
\end{table}%

\setlength{\tabcolsep}{1.75mm}
\begin{table*}[htbp]
\scriptsize
  \centering
  \caption{The evaluation of each convolutional layer for VGG-16 on CIFAR-10.}
    \begin{tabular}{|c|c|c|ccccccccccccc|}
    \hline
    \multirow{3}[0]{*}{Model} & \multirow{3}[0]{*}{Epochs} & \multirow{3}[0]{*}{Top1-Acc} & \multicolumn{13}{c|}{AFIE} \\
    \cline{4-16}
    & & &$\emph{AFIE}_{1}$&$\emph{AFIE}_{2}$&$\emph{AFIE}_{3}$&$\emph{AFIE}_{4}$&$\emph{AFIE}_{5}$&$\emph{AFIE}_{6}$&$\emph{AFIE}_{7}$&$\emph{AFIE}_{8}$&$\emph{AFIE}_{9}$&$\emph{AFIE}_{10}$&$\emph{AFIE}_{11}$&$\emph{AFIE}_{12}$&$\emph{AFIE}_{13}$\\
    \hline
    \multirow{3}[0]{*}{VGG-16}&1 & 0.435  & 0.016 & 0.064 & 0.032 & 0.038 & 0.019&0.022&0.011&0.012&0.012&0.012&0.012&0.012&0.012 \\
     \cline{2-16}
    &50 & 0.905 & 0.016 & 0.064 & 0.032 & 0.038 & 0.019&0.022&0.011&0.012&0.012&0.012&0.012&0.012&0.012 \\
    \cline{2-16}
    &150 & 0.935 & 0.016 & 0.064 & 0.032 & 0.038 & 0.019&0.022&0.011&0.012&0.012&0.012&0.012&0.012&0.012 \\
    \hline
    \end{tabular}%
  \label{k_vgg16}%
\end{table*}%

\setlength{\tabcolsep}{0.8mm}
\begin{table*}[htbp]
\scriptsize
  \centering
  \caption{The evaluation of each convolutional layer for ResNet-50 on ImageNet.}
    \begin{tabular}{|c|c|c|cccccccccccccccc|}
    \hline
    \multirow{3}[0]{*}{Model} & \multirow{3}[0]{*}{Epochs} & \multirow{3}[0]{*}{Top1-Acc} & \multicolumn{16}{c|} {AFIE} \\
    \cline{4-19}
    & & &$\emph{AFIE}_{1}$&$\emph{AFIE}_{2}$&$\emph{AFIE}_{3}$&$\emph{AFIE}_{4}$&$\emph{AFIE}_{5}$&$\emph{AFIE}_{6}$&$\emph{AFIE}_{7}$&$\emph{AFIE}_{8}$&$\emph{AFIE}_{9}$&$\emph{AFIE}_{10}$&$\emph{AFIE}_{11}$&$\emph{AFIE}_{12}$&$\emph{AFIE}_{13}$&$\emph{AFIE}_{14}$&$\emph{AFIE}_{15}$&$\emph{AFIE}_{16}$\\
    \hline
    \multirow{3}[0]{*}{ResNet-50}&1 & 0.435  & 0.064 & 0.064 & 0.064 & 0.038 & 0.038&0.038&0.038&0.022&0.022&0.022&0.022&0.022&0.022&0.012&0.012&0.012 \\
     \cline{2-19}
    &50& 0.646  & 0.064 & 0.064 & 0.064 & 0.038 & 0.038&0.038&0.038&0.022&0.022&0.022&0.022&0.022&0.022&0.012&0.012&0.012\\
    \cline{2-19}
    &150& 0.758  & 0.064 & 0.064 & 0.064 & 0.038 & 0.038&0.038&0.038&0.022&0.022&0.022&0.022&0.022&0.022&0.012&0.012&0.012\\
    \hline
    \end{tabular}%
  \label{k_resnet50}%
\end{table*}%

\begin{figure}
    \centering
    \includegraphics[width=3.3in]{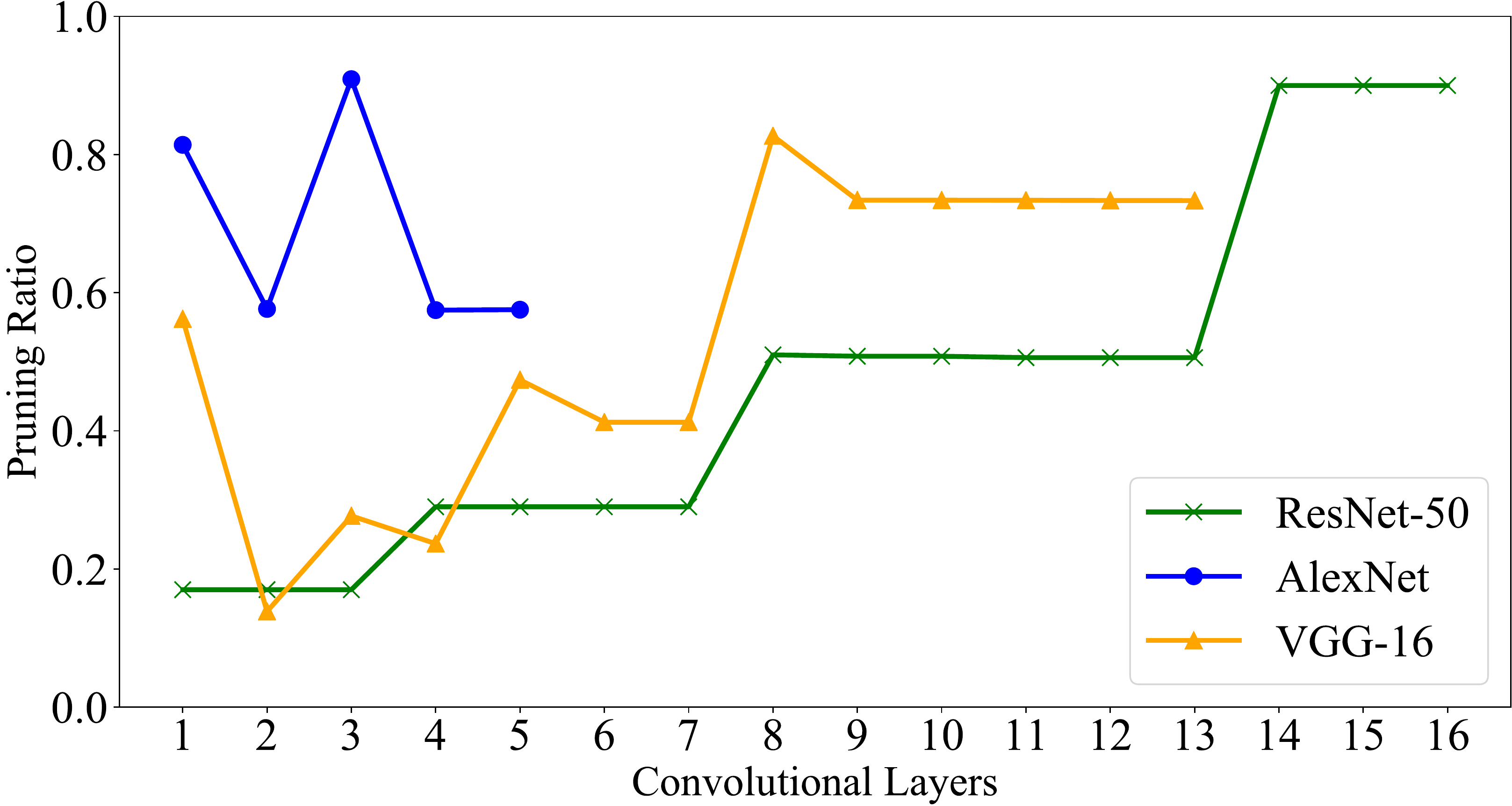}
    \caption{The pruning ratio of each convolutional layer for AlexNet, VGG-16 and ResNet-50 on MNIST, CIFAR-10 and ImageNet, respectively. }
    \label{pruning_details}
\end{figure}

\setlength{\tabcolsep}{0.9mm}
\begin{table}[htbp]
\scriptsize
  \centering
  \caption{The comparison of different pruning methods for AlexNet, VGG-16 and ResNet-50. The notations "Par-O", "FLOPs-O" and "Par-P", "FLOPs-P" stand for the parameters, and FLOPs for the original model and the slimmed model, respectively.}
    \begin{tabular}{|c|c|c|c|c|c|c|c|}
    \hline
    Model & Methods &  Par-O& FLOPs-O& Pru-R & Par-P & FLOPs-P & Top1-Acc \\
    \hline
    \multirow{5}[0]{*}{AlexNet} & ThiNet & \multirow{5}[0]{*}{3.52M} &    \multirow{5}[0]{*}{1.3$\times$$10^{7}$} &\multirow{5}[0]{*}{0.7}    & 1.10M&   1.0$\times$$10^{7}$ & 0.9920 \\
    \cline{2-2}\cline{6-8}
          & $\emph{l}_{1}$    &    & &   &    1.03M  &  1.0$\times$$10^{7}$& 0.9918 \\
           \cline{2-2}\cline{6-8}
          & Net Slim &      & & &   1.16M & 1.0$\times$$10^{7}$  &  0.9921  \\
           \cline{2-2}\cline{6-8}
          & Taylor &       &    & &   0.55M &1.0$\times$$10^{7}$ &  0.9925\\
          \cline{2-2}\cline{6-8}
          & \textbf{AFIE}  &       & & &    \textbf{0.83M} &  \textbf{1.0}\bm{$\times$}\bm{$10^{7}$}   &\textbf{0.9920}\\
          \hline
    \multirow{5}[0]{*}{VGG-16} & ThiNet &\multirow{5}[0]{*}{33.65M} &   \multirow{5}[0]{*}{3.3$\times$$10^{8}$}    &     \multirow{5}[0]{*}{0.65}   & 19.25M & 1.6$\times$$10^{8}$&0.9312\\
     \cline{2-2}\cline{6-8}
          & $\emph{l}_{1}$    &       &  & &  19.60M    &  1.6$\times$$10^{8}$     &0.9300  \\
          \cline{2-2}\cline{6-8}
          & Net Slim &       &  & &19.39M      &1.6 $\times$$10^{8}$       &  0.9335\\
           \cline{2-2}\cline{6-8}
          & Taylor &       &  &  &     18.55M  & 1.0 $\times$ $10^{8}$ & 0.9106 \\
           \cline{2-2}\cline{6-8}
          & \textbf{AFIE}  &       &  & & \textbf{19.15M}     &   \textbf{1.5}\bm{$\times$}\bm{$10^{8}$}& \textbf{0.9335}  \\
          \hline
              \multirow{5}[0]{*}{ResNet-50} & ThiNet & \multirow{5}[0]{*}{25.56M} &    \multirow{5}[0]{*}{4.12$\times$$10^{9}$}   &  \multirow{5}[0]{*}{0.3}&  21.72M& 3.56$\times$$10^{9}$&0.758  \\
     \cline{2-2}\cline{6-8}
          & $\emph{l}_{1}$    &       &    & & 22.23M  &  3.71$\times$$10^{9}$     & 0.755 \\
          \cline{2-2}\cline{6-8}
          & Net Slim &       & & &  21.95M    &   3.60$\times$$10^{9}$ &  0.758 \\
           \cline{2-2}\cline{6-8}
          & Taylor &       &     & & 20.85M &  3.15$\times$$10^{9}$     & 0.741 \\
           \cline{2-2}\cline{6-8}
          & \textbf{AFIE}  &       &     & &   \textbf{21.88M}&  \textbf{3.7}\bm{$\times$}\bm{$10^{9}$}     & \textbf{0.758} \\
          \hline
    \end{tabular}%
  \label{pruning_results}%
\end{table}%

\subsection{Implementation of Filters Removal}
After obtaining AFIE scores for all the layers, we randomly remove filters within each layer, under the guidance of these AFIE scores. There are two categories of pruning strategies for removing filters, as follows.

Iterative pruning strategy removes a part of the filters in several iterations. This requires dynamic evaluation of AFIE. The whole pruning process continues until the overall pruning ratio reaches to the specified $\lambda^{*}$. However, iterative pruning may lead the pruned model to be trapped into the local optima, since AFIE scores need to be re-calculated during each iteration.

One-shot pruning strategy removes all the filters once from all layers. Therefore, AFIE scores only need to be calculated for one time. Obviously, one-shot pruning avoids the re-calculation of AFIE for each convolutional layer, which can improve the efficiency of filter removal, as well as skipping the trap of local optima. Therefore, in the experiments, we adopt one-shot pruning.

\subsection{Overall Evaluation}
We implement our pruning framework AFIE based on AlexNet, VGG-16, and ResNet-50, and test them on MNIST, CIFAR-10, and ImageNet.

First, we show the AFIE scores of each convolutional layer, for models that are trained with different epochs. The results are shown in Table \ref{k_alexnet}, Table \ref{k_vgg16} and Table \ref{k_resnet50}. As we can see, the AFIE stay unchanged no matter how many iterations the original model is trained. Even when the original model is trained for only one epoch, our method can still learn effective AFIE scores. Therefore, we can use the proposed AFIE to evaluate the importance of each filter without considering the interference of weight information.

Then, we compare the performance with several previous structured pruning methods. The result is shown in Table \ref{pruning_results}. As we can see, our AFIE method can recover the accuracy of baselines on all the datasets. Particularly, AFIE is able to reduce the parameters and FLOPs than all the baselines except for the Taylor baseline. Nevertheless, Taylor typically sacrifices more accuracy. Therefore, our AFIE is a competitive method for the network pruning.

% For AlexNet on MNIST, our method can recover baseline accuracy, as well as achieve same reduction of FLOPs as other methods, and the parameter reduction can outperform other methods exclude Taylor.

% For VGG-16, only our method and "Net Slim" can recover baseline accuracy, but AFIE method can yield better compression of parameters and FLOPs than "Net Slim".

\subsection{Filters Pruning for AlexNet on MNIST}
AlexNet is constructed by 5 convolutional layers with BachNorm embedded. As shown in Table \ref{k_alexnet}, the original model can achieve $99.20\%$ top1-accuracy with 20 training epochs on MNIST. We set the overall pruning ratio $\lambda^{*}$ to $70\%$, and specify the specific pruning ratio each layer according to the Equation (\ref{sprasity_calculation}). To maximally demonstrate the good consistency of AFIE, we prune AlexNet when it is trained with only one epoch. 

Figure \ref{pruning_details} illustrates the details of filters pruning. As shown, Conv2 is assigned with the highest pruning ration, i.e., $57.65\%$. The pruning ratios of Conv1, Conv3, Conv4 and Conv5 are $81.4\%$, $90.91\%$, $57.48\%$, and $57.53\%$, respectively. According to these pruning ratios, we prune each layer of the original model in a one-shot way. The pruned results are displayed in Table \ref{pruning_results}. Obviously, our pruning framework can achieve $76.42\%$ and $23.08\%$ reduction both on parameter and FLOPs, as well as maintain a strong representation ability. Therefore, our AFIE is an efficient criteria to help slim the under-trained AlexNet on MNIST.

\subsection{Filters Pruning for VGG-16 on CIFAR-10} 
VGG-16 inherits the chain structure of AlexNet while extends the number of convolutional layers to 13. The original model achieved $93.35\%$ top1-accuracy on CIFAR-10 with 150 training epochs. Similar to the previous experiment, we show the pruning ratio of each layer in Figure \ref{pruning_details}. As we can see, the overall pruning ratio $\lambda^{*}$ is set as $65\%$, and Conv2 has the lowest pruning ratio with $13.84\%$. Conv1 is allotted with $56.19\%$ pruning ratio. Conv3, Conv4 and Conv5 are allocated with $27.68\%$, $23.66\%$ and $47.37\%$ pruning ratios, respectively. Conv6 and Conv7 are pruned with $41.24\%$ ratio. Finally, the remaining layers are assigned with highest pruning ratio around $73.40\%$.  

Then, one-shot pruning is conducted to prune the original model with the obtained pruning ratios. The results are displayed on Table \ref{pruning_results}. Clearly, our AFIE pruning framework achieves impressive results with $43.09\%$ and $54.55\%$ reduction for parameters and FLOPs respectively, and outperforms previous pruning methods with higher accuracy. Note that our AFIE can achieve such performance even the original model is trained for only one epoch, while baseline method need to fully pre-train the original model and require iterative training process.

\begin{figure*}[ht]
 \subfigure[Conv8]{
        \begin{minipage}[t]{0.32\linewidth}
          \centering
          \includegraphics[width=2.3in]{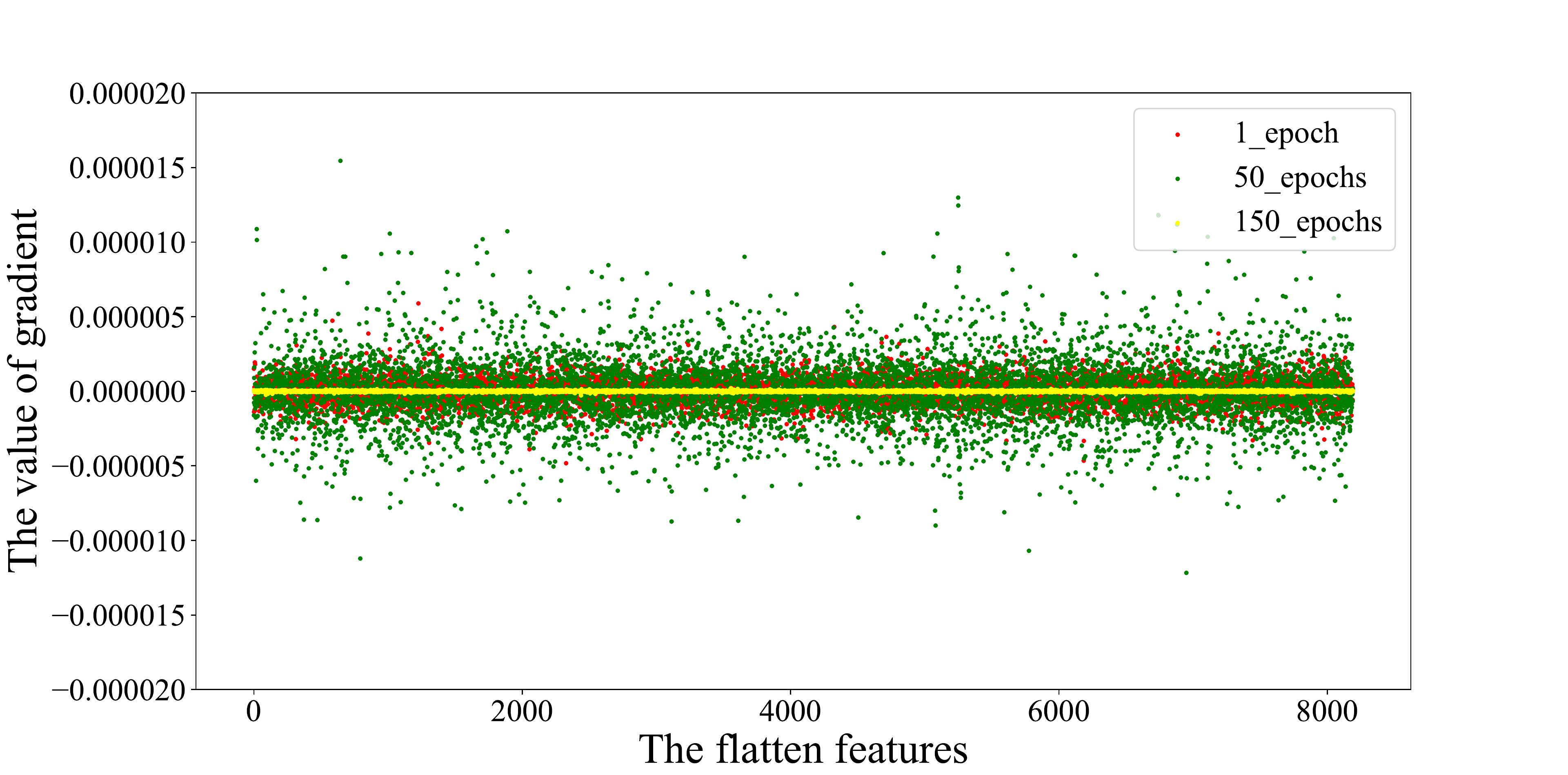}
      %\caption{fig1}
        \end{minipage}
        }
      \subfigure[Conv9]{
        \begin{minipage}[t]{0.32\linewidth}
          \centering
          \includegraphics[width=2.3in]{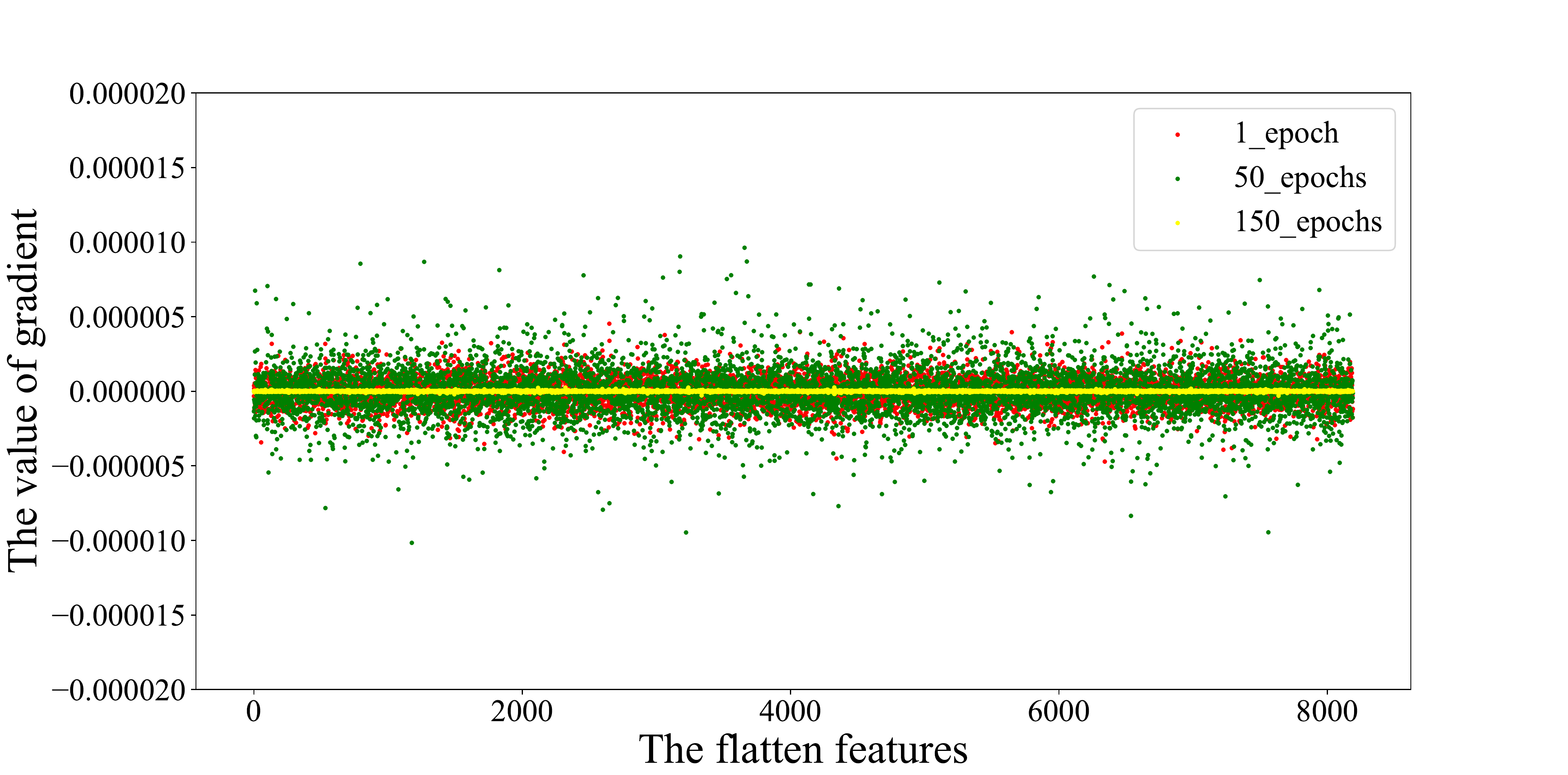}
          %\caption{fig2}
        \end{minipage}
      }
       \subfigure[Conv10]{
        \begin{minipage}[t]{0.32\linewidth}
          \centering
          \includegraphics[width=2.3in]{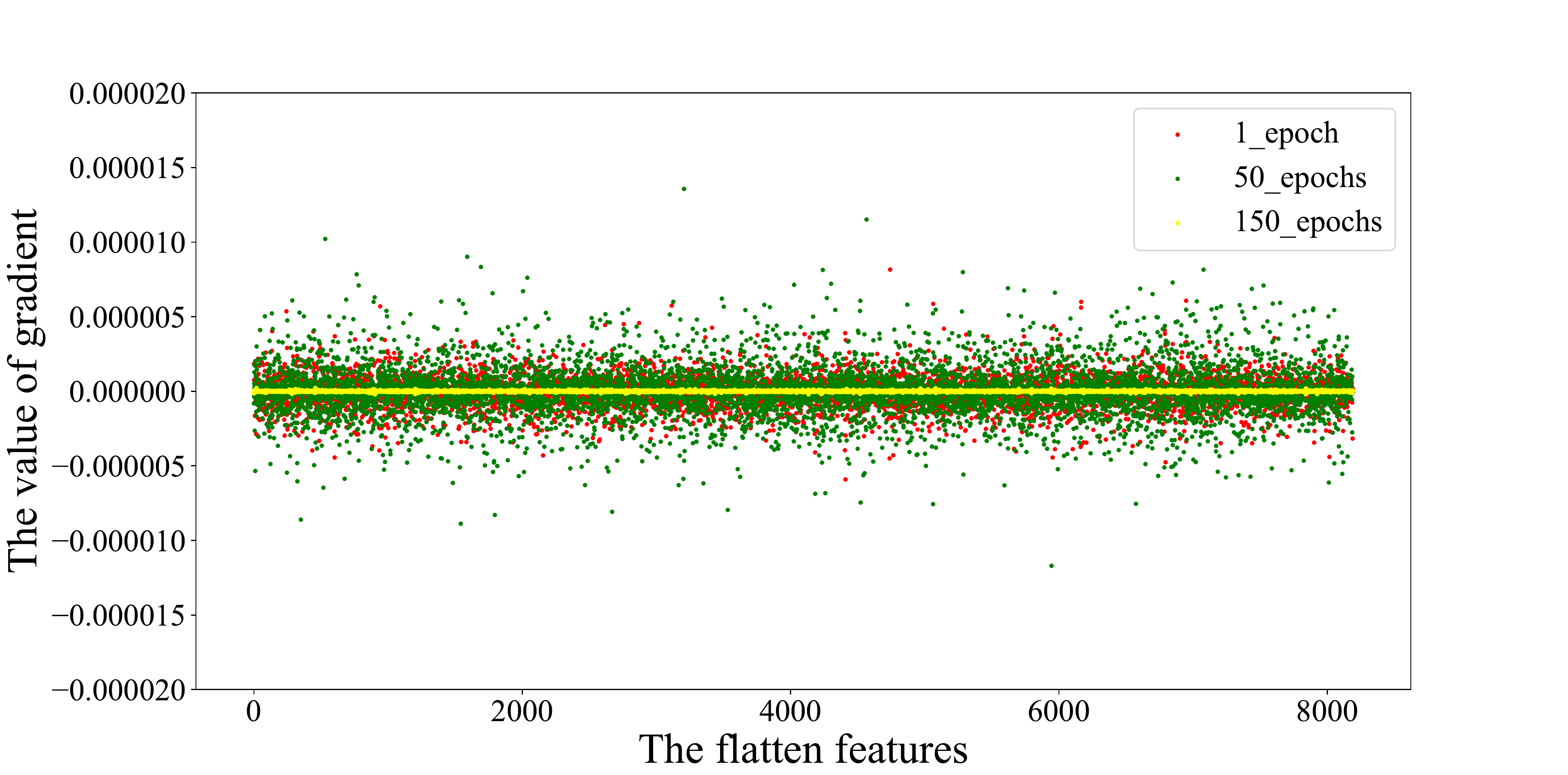}
      %\caption{fig1}
        \end{minipage}
        }
        
      \subfigure[Conv11]{
        \begin{minipage}[t]{0.32\linewidth}
          \centering
          \includegraphics[width=2.3in]{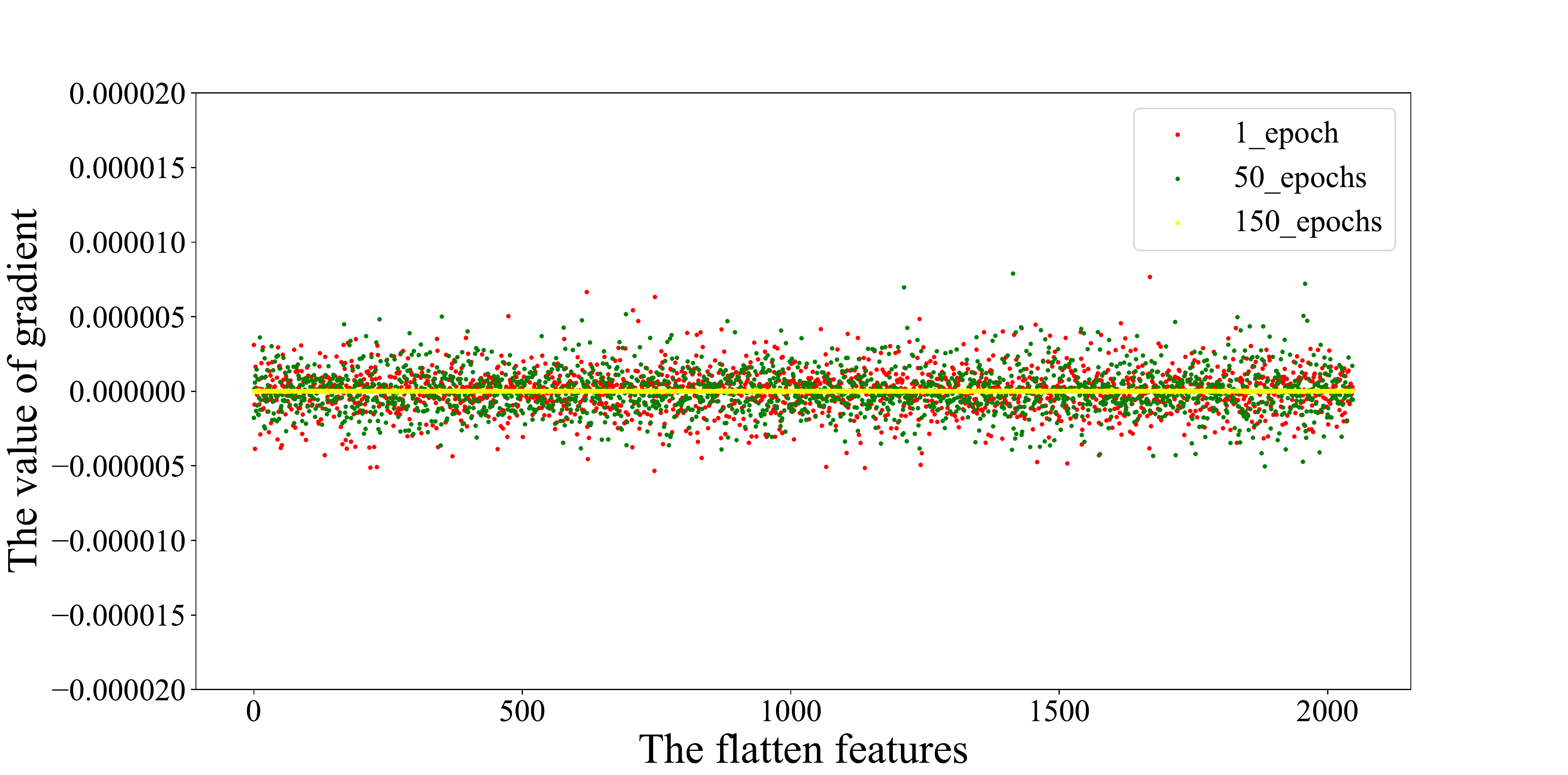}
          %\caption{fig2}
        \end{minipage}
      }
         \subfigure[Conv12]{
        \begin{minipage}[t]{0.32\linewidth}
          \centering
          \includegraphics[width=2.3in]{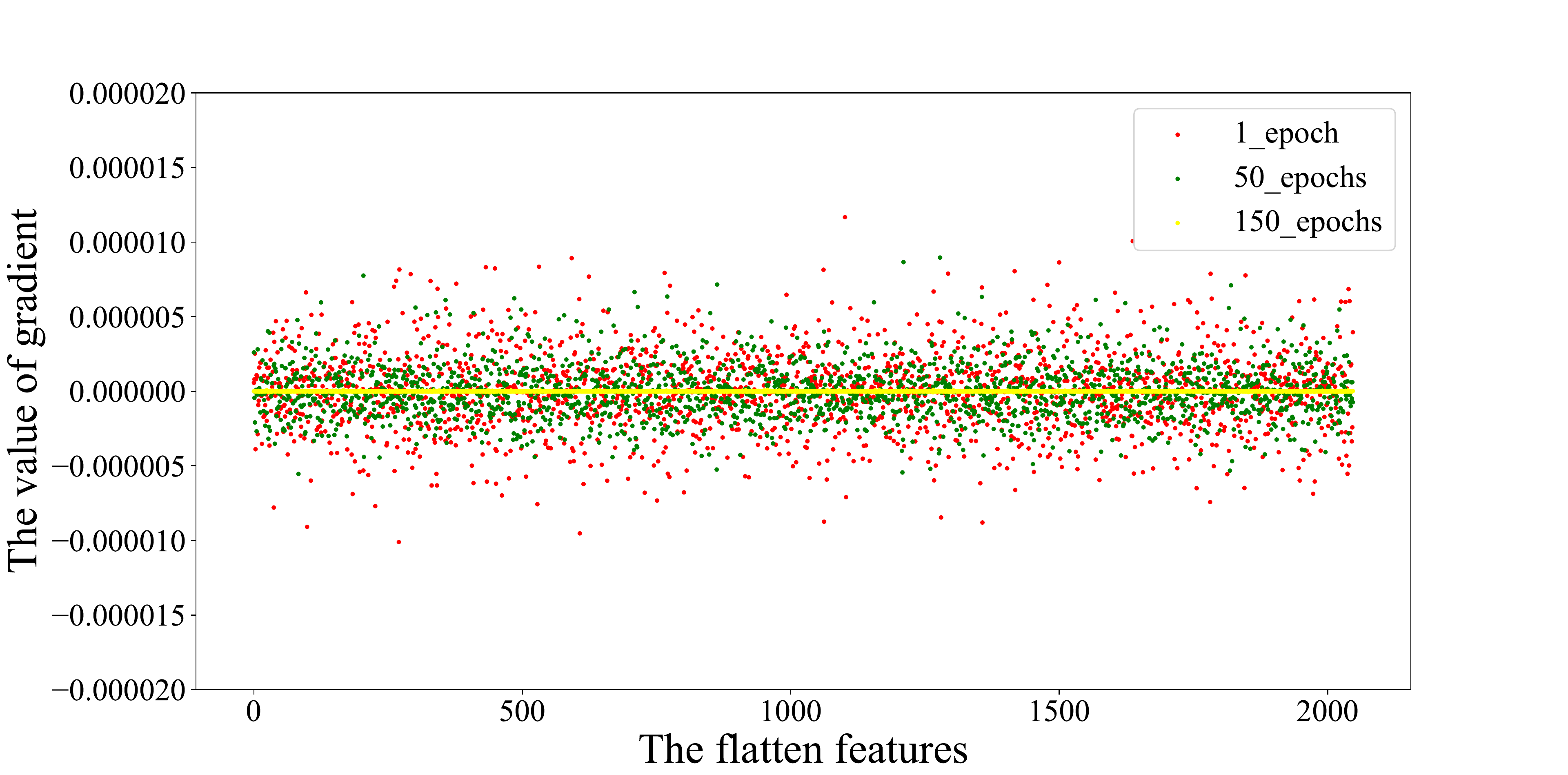}
      %\caption{fig1}
        \end{minipage}
        }
      \subfigure[Conv13]{
        \begin{minipage}[t]{0.32\linewidth}
          \centering
          \includegraphics[width=2.3in]{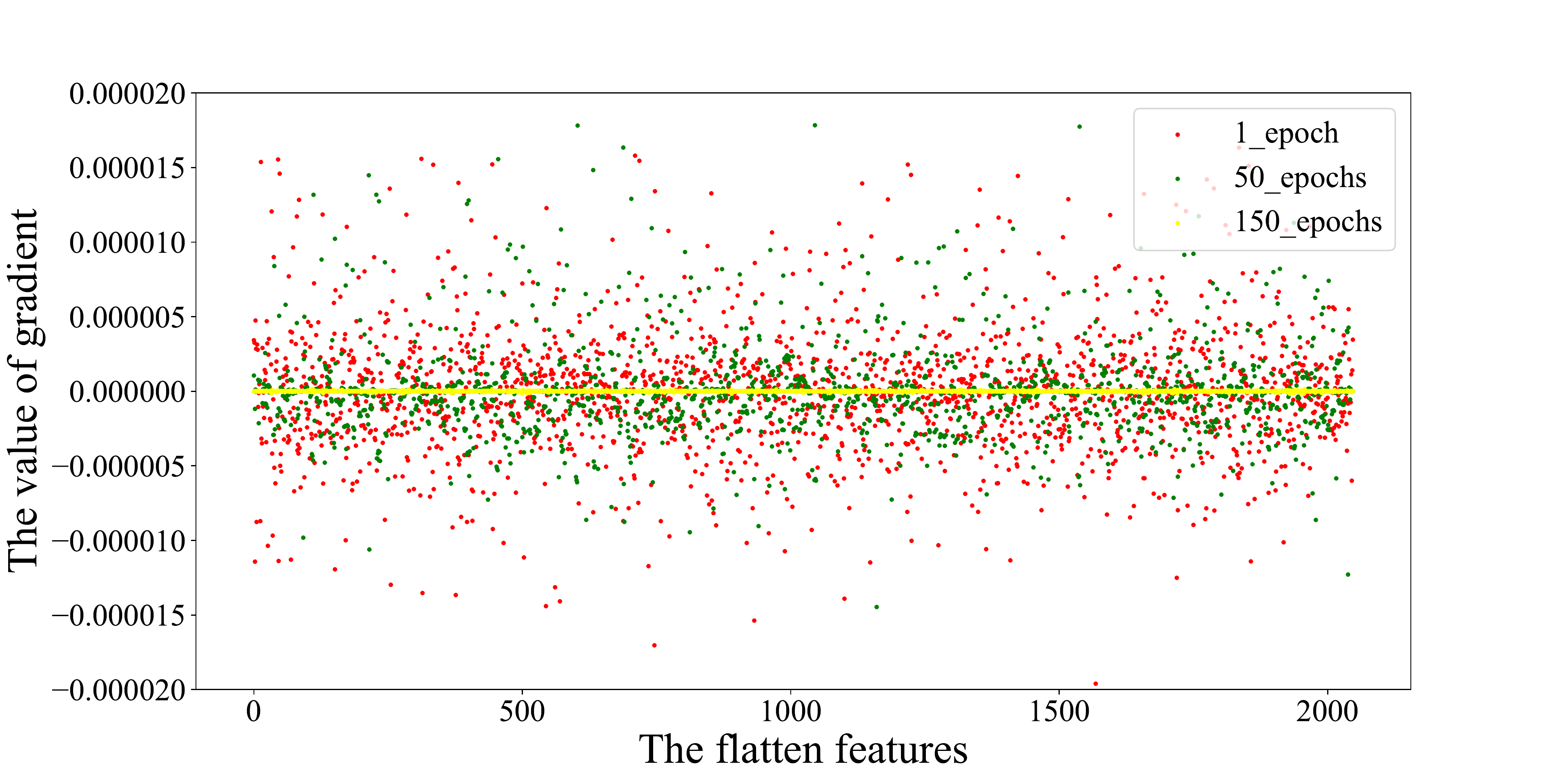}
          %\caption{fig2}
        \end{minipage}
      }
    \caption{The distribution of flatten output features for each convolutional layer. The x-axis and y-axis stand for the index and value of each flattened feature point. The red, green and yellow points stand for the flattened feature points for the models that have been trained 1, 50 and 150 epochs, respectively.}
      \label{flatten_points}
    \end{figure*}

\subsection{Filters Pruning for ResNet-50 on ImageNet}
Through the above experiments, we have shown the superiority of our AFIE pruning framework on relatively shallow and deep chain structure networks. Here, we further extend the work on ResNet-50 with more complicated structures. ResNet-50 is designed with 4 blocks and multiple skip connections. It has better representation ability on larger dataset like ImageNet. For this model, we only remove the filters of the second convolutional layers within each block with the size of $3\times3$. For simplicity, we rename the corresponding layers from Conv1 to Conv16. Notably, the original model can achieve $30.8\%$ and $75.8\%$ top1-accuracy with 1 and 150 training epochs, respectively. 

Then, we set the overall pruning ratio $\lambda^{*}$ to $30\%$, and specify the pruning ratios of Conv1, Conv2, and Conv3 to $17\%$, Conv4, Conv5, Conv6, and Conv7 to $29\%$, Conv8, Conv9, Conv10, Conv11, Conv12, and Conv13 $51\%$, and Conv14, Conv15, and Conv16 to $90\%$. As shown in Figure \ref{pruning_details}, we can see that the distribution of the pruning ratios within each block is flat, which means that the filters within same block have similar importance. We perform one-shot pruning according to these ratios. The results are listed on Table \ref{pruning_results}. As shown, our method can still achieve comparable pruning results as other baselines.

\subsection{Gradient Visualization for Output Features}
In this section, we visualize the distribution of the gradient for the output features of each convolutional layer. This can help illustrate that the updating of parameters between the well-trained and under-trained models is lipchitz continuous. Specifically, we plot the gradient of VGG-16 when it is trained with 1, 50 and 150 epochs, respectively. The bath-size of the training set for CIFAR-10 is set to 128, and the learning rate is set to [0.01, 0.005, 0.001, 0.0005, 0.0001] for every 30 epochs. For each convolutional layer, the gradient of the features map is recorded after feeding of the data. The gradient tensor for the output features can be denoted as $\textbf{G}^{B\times O\times W_{r}\times H_{r}}$, where $B$ is the size of batch for input data, $O$ is the number of output features, and $W_{r}$ and $H_{r}$ stand for the reduced width and height of each feature map after pooling. Especially, we get the final gradient of each feature map by averaging all the batches and flattening the feature maps as a vector $\textbf{g}^{1\times (O\times W_{r}\times H_{r})}$. 

The visualization results for VGG-16 on CIFAR-10 are showed as Figure \ref{flatten_points}. Since there are too many feature points in the shallow layers after flattening, we only plot the gradient of the feature maps for Conv8, Conv9, Conv10, Conv11, Conv12, and Conv13. For the last three convolutional layers, there are only around 2000 feature points. We observe that the gradient of each feature point is quite small only when the model is trained for only one epoch. Moreover, the gradient of all feature points are distributed within the magnitude between $-10^{-5}\sim10^{-5}$. It means that the updating of the parameters for the Conv11, Conv12, and Conv13 are small between the well-trained and under-trained models. Similarly, the values of the feature points for Conv8, Conv9, and Conv10 are also distributed between $-10^{-5}\sim 10^{-5}$, with more narrow variation for the above three models. It implies that the corresponding updating of parameters are small when the model is trained with one epoch, verifying our theoretical analysis in Remark 1.

\section{Conclusion}
In this paper, we propose an entropy based pruning framework AFIE to evaluate the importance of each filter. We verify its pruning effectiveness for models of AlexNet, VGG-16 and ResNet-50, on datasets of MNIST, CIFAR-10 and ImageNet, respectively. We analyze that AFIE can stay stable both for the fully-trained and under-trained model by the Lipschitz limitation parameters. It means that the evaluation of AFIE is parameter-free, therefore, we can get a well slimmed model even when the original model is only trained with one epoch.

\bibstyle{aaai23}
\bibliography{aaai23}
\end{document}